\documentclass[10pt,twocolumn,letterpaper]{article}

\usepackage{cvpr}              

\usepackage{amsmath, bm}
\usepackage{amssymb}
\usepackage{booktabs}

\usepackage{algorithm,algpseudocode}
\usepackage{algorithmicx}
\usepackage{xcolor}  
\usepackage{tabularx}
\usepackage{longtable}
\usepackage{arydshln}
\usepackage{array}    
\usepackage{cuted}
\usepackage{float}

\definecolor{my_green}{RGB}{51,102,0}
\definecolor{my_red}{RGB}{204, 0, 0}
\definecolor{my_orange}{RGB}{240, 188, 124}
\definecolor{my_light_blue}{RGB}{181, 212, 216}
\newcommand{\cmark}{\textcolor{my_green}{\ding{51}}} 
\newcommand{\xmark}{\textcolor{my_red}{\ding{55}}} 
\newcommand{\datasetname}{Video-MME\xspace}

\definecolor{cvprblue}{rgb}{0.21,0.49,0.74}
\usepackage[pagebackref,breaklinks,colorlinks,allcolors=cvprblue]{hyperref}

\def\paperID{5416} 
\def\confName{CVPR}
\def\confYear{2025}

\title{Video-MME: The First-Ever Comprehensive Evaluation Benchmark \\ of Multi-modal LLMs in Video Analysis}

\author{
    Chaoyou Fu$^{1,2,\spadesuit}$, Yuhan Dai$^{3}$, Yongdong Luo$^{4}$, Lei Li$^{5}$, Shuhuai Ren$^{6}$ \\ 
    Renrui Zhang$^{7}$, Zihan Wang$^{8}$, Chenyu Zhou$^{4}$, Yunhang Shen$^{4}$, Mengdan Zhang$^{9}$ \\ 
    Peixian Chen$^{4}$, Yanwei Li$^{7}$, Shaohui Lin$^{8}$, Sirui Zhao$^{3}$, Ke Li$^{4}$, Tong Xu$^{3}$ \\ 
    Xiawu Zheng$^{4}$, Enhong Chen$^{3}$, Caifeng Shan$^{1,2,*}$, Ran He$^{9,*}$, Xing Sun$^{5,}$\thanks{Corresponding Author. $^{\spadesuit}$~Project Leader.}  \\
    \\
    $^{1}$State Key Laboratory for Novel Software Technology, Nanjing University \\
    $^{2}$School of Intelligence Science and Technology, Nanjing University \\
    $^{3}$State Key Laboratory of Cognitive Intelligence \\
    $^{4}$XMU, $^{5}$HKU, $^{6}$PKU, $^{7}$CUHK, $^{8}$ECNU, $^{9}$CASIA
}

\usepackage{adjustbox}
\usepackage{multirow}
\usepackage{pifont}
\usepackage{wrapfig}
\usepackage{framed}
\usepackage{makecell}
\usepackage{listings}
\usepackage{csquotes}

\begin{document}
\maketitle

\begin{abstract}

In the quest for artificial general intelligence, Multi-modal Large Language Models (MLLMs) have emerged as a focal point in recent advancements. 
However, the predominant focus remains on developing their capabilities in static image understanding.
The potential of MLLMs to process sequential visual data is still insufficiently explored, highlighting the lack of a comprehensive, high-quality assessment of their performance. In this paper, we introduce \textbf{Video-MME}, the first-ever full-spectrum, \textbf{M}ulti-\textbf{M}odal \textbf{E}valuation benchmark of MLLMs in \textbf{Video} analysis.
Our work distinguishes from existing benchmarks through four key features: 
\textit{\textbf{1) Diversity in video types}}, spanning 6 primary visual domains with 30 subfields to ensure broad scenario generalizability;
\textit{\textbf{2) Duration in temporal dimension}}, encompassing both short-, medium-, and long-term videos, ranging from 11 seconds to 1 hour, for robust contextual dynamics;
\textit{\textbf{3) Breadth in data modalities}}, integrating multi-modal inputs besides video frames, including subtitles and audios, to unveil the all-round capabilities of MLLMs;
\textit{\textbf{4) Quality in annotations}}, utilizing rigorous manual labeling by expert annotators to facilitate precise and reliable model assessment. 
With Video-MME, we extensively evaluate various state-of-the-art MLLMs, and reveal that Gemini 1.5 Pro is the best-performing commercial model, significantly outperforming the open-source models with an average accuracy of 75\%, compared to 71.9\% for GPT-4o. 
The results also demonstrate that \datasetname is a universal benchmark that applies to both image and video MLLMs.
Further analysis indicates that subtitle and audio information could significantly enhance video understanding. 
Besides, a decline in MLLM performance is observed as video duration increases for all models.
Our dataset along with these findings underscores the need for further improvements in handling longer sequences and multi-modal data, shedding light on future MLLM development.
Project page: \url{https://video-mme.github.io}.
\end{abstract}
    
\section{Introduction}
\begin{figure*}
    \centering
    \includegraphics[width=0.89\textwidth]{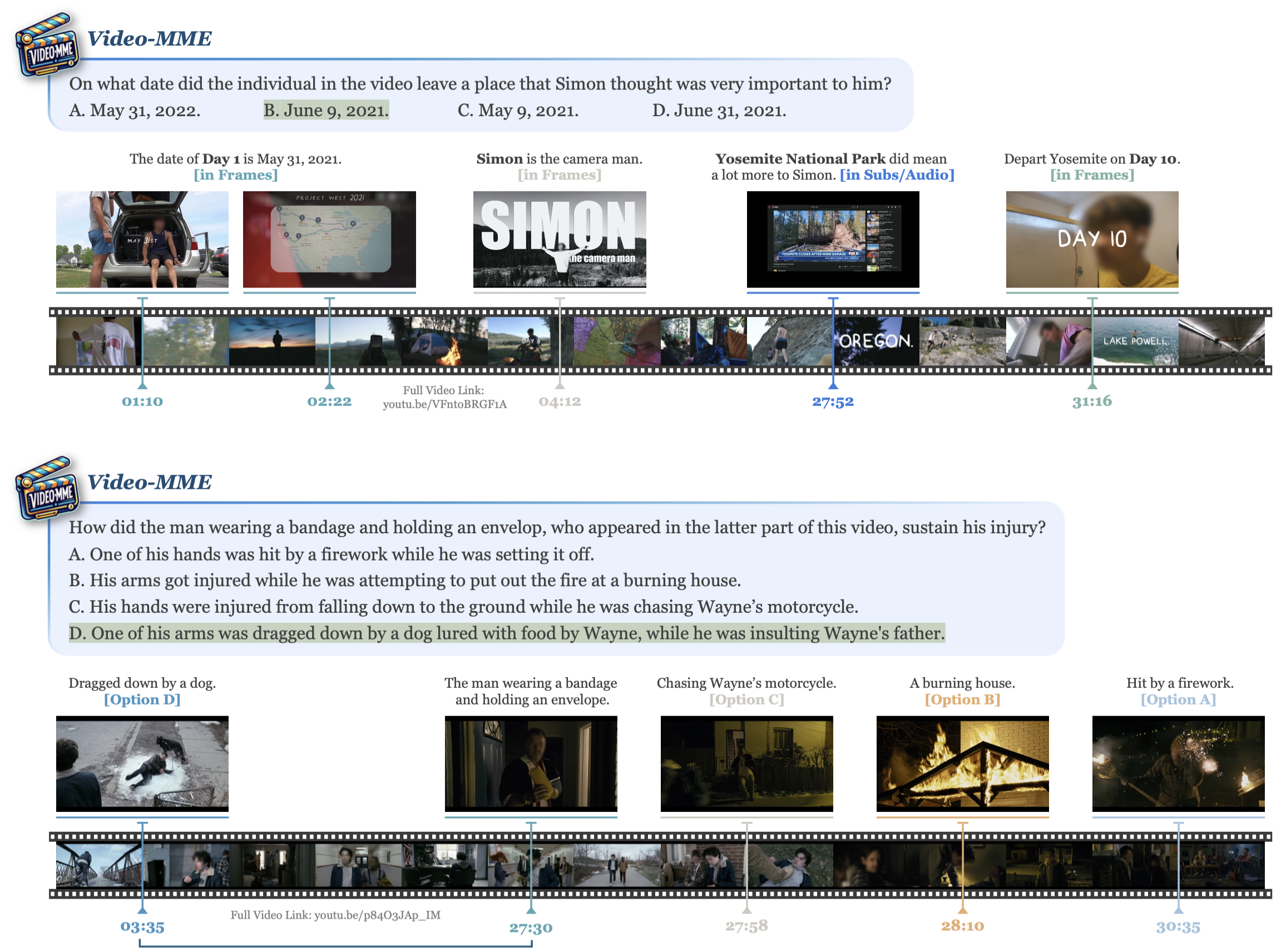}
    \captionof{figure}{We introduce Video-MME to provide high-quality assessment of MLLMs' performance, where all the videos and annotations are \textbf{manually} collected and curated.
    These are two highlighted examples of Video-MME with the ground-truth answers in \textcolor{my_green}{green}, which require the model to perform long-term contextual understanding and complex spatial and temperal reasoning.
    Video-MME is meticulously designed to pose significant challenges, thereby effectively evaluating the compositional video understanding capabilities of MLLMs.}
    \label{fig:video-mme-case}
\end{figure*}
\begin{table*}[ht]
\centering

\begin{adjustbox}{max width=0.9\textwidth}
\begin{tabular}{lrrrrcrccccc}
\toprule
\textbf{Benchmarks} & \textbf{\#Videos} & \textbf{\#Clips} &\textbf{Len.(s)} & \textbf{\#QA Pairs} &  \textbf{Anno.} & \textbf{QA Tokens} & \textbf{Sub. Tokens} & \textbf{Multi-level}  & \textbf{Open-domain} & \textbf{Sub.\&Aud.} \\
\midrule
MSRVTT-QA~\citep{Xu2017VideoQA} & 2,990 & 2,990 & 15.2 & 72,821 & A & 8.4 & \xmark & \xmark & \cmark & \xmark\\
MSVD-QA~\citep{Xu2017VideoQA}& 504 & 504 & 9.8 & 13,157 & A & 7.6 & \xmark & \xmark & \cmark & \xmark \\
TGIF-QA~\citep{jang2017tgif} & 9,575 & 9,575 & 3.0 & 8,506 & A\&M & 20.5 & \xmark & \xmark & \cmark & \xmark \\
ActivityNet-QA \citep{yu2019activitynet} & 800 & 800 & 111.4 & 8,000 & M & 10.2 & \xmark & \xmark & \xmark & \xmark\\
TVQA~\citep{lei2018tvqa} & 2,179 & 15,253 & 11.2 & 15,253 & M & 27.8 & 159.8 & \xmark & \xmark & \xmark\\
How2QA~\citep{li2020hero}& 1,166 & 2,852 & 15.3 & 2,852 & M & 16.9 & 31.1 & \xmark & \cmark & \xmark \\
STAR~\citep{wu2021star_situated_reasoning} & 914 & 7,098 & 11.9 & 7,098 & A & 19.5 & \xmark & \xmark & \cmark & \xmark\\
NExT-QA~\citep{xiao2021next} & 1,000 & 1,000 & 39.5 & 8,564 & A & 25.3 & \xmark & \xmark & \cmark & \xmark \\
\midrule
MVBench~\citep{li2023mvbench} & 3,641 & 3,641 & 16.0 & 4,000 & A & 27.3 & \xmark & \xmark & \cmark & \xmark\\
Video-Bench~\citep{ning2023video} & 5,917 & 5,917 & 56.0 & 17,036 & A\&M & 21.3 & \xmark & \xmark & \cmark & \xmark\\
EgoSchema~\citep{mangalam2024egoschema} & 5,063 & 5,063 & 180.0 & 5,063 & A\&M & 126.8 & \xmark & \xmark & \xmark & \xmark \\
AutoEval-Video~\citep{chen2023autoeval} & 327 & 327 & 14.6 & 327 & M & 11.9 & \xmark & \xmark & \cmark & \xmark\\
TempCompass~\citep{Liu2024TempCompassDV} & 410 & 500 & 11.4 & 7,540 & A\&M & 49.2 & \xmark & \xmark & \cmark & \xmark\\
\midrule
\textbf{Video-MME-S} & 300 & 300 & 80.7 & 900 &  \multirow{4}{*}{M} & 28.7 & 198.6 & \multirow{4}{*}{\textcolor{green}\cmark} & \multirow{4}{*}{\cmark} &  \multirow{4}{*}{\cmark} \\
\textbf{Video-MME-M} & 300 & 300 & 515.9 & 900 &  & 32.8 & 1425.6 &     \\
\textbf{Video-MME-L} & 300 & 300 & 2466.7 & 900 &  & 45.6 & 6515.6 &     \\
\textbf{Video-MME} & 900 & 900 & 1017.9 & 2,700 &  & 35.7 & 3086.5 &     \\
\bottomrule
\end{tabular}
\end{adjustbox}
\caption{\textbf{The comparison of various benchmarks} encompasses several key aspects: the total number of videos (\textbf{\#Videos}), the number of clips (\textbf{\#Clips}), the average duration of the videos (\textbf{Len.}), the number of QA pairs (\textbf{\#QA Pairs}), the method of annotation (\textbf{Anno.}, M/A means the manually/automatic manner), the average number of QA pair tokens (\textbf{QA Tokens}), the average number of subtitle tokens (\textbf{Sub. Tokens}), whether the videos cover multiple duration levels (\textbf{Multi-level}), whether the videos are sourced from a broad range of open domains (\textbf{Open-domain}), and whether provide subtitle together with audio information (\textbf{Sub.\&Aud.}). Video-MME-S/M/L denotes the short/medium/long part. It is important to note our comparison
focuses solely on the multiple-choice subset of the datasets.} 

\label{tab:comparison}
\vspace{-1em}
\end{table*}

The rapid development of MLLMs in recent years \citep{yin2023survey} has highlighted their impressive perception and cognitive capabilities across various multimodal benchmarks \citep{fu2023mme,Yu2023MMVetEL,Lu2023MathVistaEM}. 
These advancements show the great potential of MLLMs to serve as a foundation that can digest the multimodal real world and pave the way toward artificial general intelligence.
However, current MLLMs and their evaluation primarily focus on static visual data understanding, which fails to capture the dynamic nature of the real world involving complex interactions between objects over time. 
To approximate real-world scenarios more accurately, it is crucial to explore and assess the capabilities of MLLMs on sequential visual data, such as videos.
Many early efforts~\citep{Reid2024Gemini1U,lin2023video,fu2025vita} have been made to inspire the video understanding potentials of MLLMs with promising results.
However, existing video-based benchmarks~\citep{li2023vitatecs,li2023mvbench,mangalam2024egoschema,Liu2024TempCompassDV} are still limited to thoroughly reveal their performance, such as a lack of diversity in video types, insufficient coverage of temporal dynamics, and the narrow focus on a single modality. 
These inevitably hinder the all-around evaluation of MLLMs.

To this end, we introduce \textbf{Video-MME}, the first-ever comprehensive \textbf{M}ulti-\textbf{M}odal \textbf{E}valuation benchmark crafted for MLLMs in \textbf{Video} analysis. As exemplified in Figure~\ref{fig:video-mme-case}, we meticulously curate a dataset of 900 videos across various scenarios, and annotate a set of 2,700 high-quality multiple-choice questions (3 per video) to foster a robust evaluation.
As presented in Figure~\ref{fig:domain_category}, for generalizability, our dataset widely spans 6 visual domains, including Knowledge, Film \& Television, Sports Competition, Artistic Performance, Life Record, and Multilingual, with 30 fine-grained categories, e.g., technology, documentary, news report, esports, magic show, and fashion. 
Importantly, the videos vary significantly in length, ranging from 11 seconds to 1 hour, specifically evaluating the adaptability of MLLMs across varying temporal contexts.
Furthermore, Video-MME enriches the assessment by incorporating the associated subtitles and audio tracks, thereby enhancing the analysis of multimodal inputs for video understanding.

Using Video-MME, we benchmark various state-of-the-art MLLMs, including GPT-4V~\citep{openai2023gpt4v}, GPT-4o~\citep{openai2024gpt4o}, and Gemini 1.5 Pro~\citep{Reid2024Gemini1U}, alongside open-source image models like InternVL-Chat-V1.5~\citep{chen2024far} and video models like LLaVA-NeXT-Video~\citep{zhang2024llavanextvideo}.
Our experiments in Table~\ref{tab:open-source-eval-results-wrt-duration} indicate that Gemini 1.5 Pro is the highest-performing commercial model, achieving an average accuracy of 75\%.
In comparison, open-source MLLMs exhibit substantial gaps compared to commercial models. 
For instance, the leading open-source model, VILA-1.5~\citep{lin2023vila}, attains an overall accuracy of 59\%. 
These findings suggest there is considerable room for improvement in the open-source community.
Our benchmark is also available to advanced image-based models by extending their input to multi-frame images, e.g., Qwen-VL-Max~\citep{Qwen-VL} and InternVL-Chat-V1.5~\citep{chen2024far}.
The accuracies of both the models reach 50\% , which is close to that of the video-specific model LLaVA-NeXT-Video, indicating that image understanding is the basis of video understanding and the wide applicability of Video-MME in the field of MLLMs.
Further observations in Table~\ref{gemini-eval-results-wrt-duration-topic} indicate that integrating subtitles and audios significantly enhances video comprehension capabilities, e.g., boosting Gemini 1.5 Pro by 6.2\% and 4.3\% respectively, with the gains being more pronounced for longer videos. 
A fine-grained analysis of task types reveals that subtitles and audios are particularly beneficial for videos requiring substantial domain knowledge.
We also note a general decline in MLLM performance with increasing video length. This trend suggests that limitations in processing longer video sequences could be a critical bottleneck in the performance of MLLMs.

Finally, we discuss promising avenues for improving the capabilities of MLLMs in processing video content. Potential directions include architectural development for better handling long context inputs and constructing training data focused on complex temporal reasoning scenarios. We expect that our benchmarking, evaluation findings, detailed analysis, and outlined insights will inspire future progress toward more capable and robust MLLMs.

\section{Related Work}

\paragraph{Advancements in MLLMs.}
Recent advancements in MLLMs have seen notable progress~\citep{yin2023survey,gao2024mini, jiang2024effectiveness,wang2023largemir,chen2023vlp}. MLLMs typically comprise three core modules: (i) a vision encoder for visual feature extraction, (ii) a modality alignment module to integrate visual features into the embedding space of the language model, and (iii) an LLM backbone for decoding multi-modal context. 
CLIP~\citep{radford2021clip} and SigLIP~\citep{Zhai2023SigmoidLF} are widely-used for image encoding, while LLaMA~\citep{touvron2023llama} and Vicuna~\citep{vicuna2023} serve as popular choices for LLMs. 
The alignment module varies from simple linear projections~\citep{liu2023llava} to more complex architectures such as Q-Former~\citep{Li2023BLIP2BL,dai2023instructblip}, and gated cross-attention layers substantiated by Flamingo~\citep{Alayrac2022FlamingoAV,awadalla2023openflamingo}. 
Additionally, Fuyu-8B~\citep{fuyu-8b} introduces a novel framework mapping raw image pixels directly to the LLM embedding space.
Regarding MLLMs for processing videos~\citep{li2023videochat, Luo2023ValleyVA, Song2023MovieChatFD}, the key difference lies in how they encode the video into vision tokens compatible with the LLMs.
Representative work like Video-LLaMA~\citep{damonlpsg2023videollama} first uses a ViT~\citep{Dosovitskiy2020AnII} with an image Q-Former to encode individual frames and then employs a video Q-Former for temporal modeling. VideoChat2~\citep{li2023mvbench} utilizes a video transformer to encode video features and subsequently implements a Q-Former~\citep{Li2023BLIP2BL} to compress video tokens. To empower video MLLMs with temporal localization capability~\citep{Huang2024LITALI, Qian2024MomentorAV, Wang2024HawkEyeTV}, TimeChat~\citep{timechat} constructs time-sensitive instruction tuning datasets and encodes timestamp knowledge into visual tokens. 
However, the potential of MLLMs in processing sequential visual data is still under-explored. 

\paragraph{MLLM Benchmarks.}
Alongside advancements in architecture, significant efforts have been made to improve benchmarking for MLLMs, guiding the development of the next generation of these models. Previous studies have integrated various aspects of evaluation, such as perception and cognitive capabilities, to create comprehensive benchmarks for assessing image MLLMs~\citep{fu2023mme,Yu2023MMVetEL,liu2023mmbench}. As image MLLMs have demonstrated exceptional performance in general perception tasks, benchmarks regarding scientific understanding~\citep{Li2024MultimodalAA}, multi-modal mathematical reasoning~\citep{Lu2023MathVistaEM,zhang2024mathverse}, and multi-disciplinary~\citep{mmmu} capabilities have drawn increasing attention.
For video MLLMs, similar efforts have been made to incorporate existing benchmarks~\citep{wang2019vatex,li2021value} for evaluating video understanding~\citep{li2023mvbench,ning2023video}. 
Due to the inherently temporal characteristics of video modalities, specialized benchmarks have been created to evaluate temporal comprehension, underscoring the current limitations of video MLLMs in understanding video content~\citep{li2023vitatecs,Liu2024TempCompassDV, wu2024longvideobenchbenchmarklongcontextinterleaved,wang2024lvbenchextremelongvideo,fang2024mmbenchvideolongformmultishotbenchmark,ataallah2024infinibenchcomprehensivebenchmarklarge}. 


\section{Video-MME}

\begin{figure*}[ht]
    \centering
    \includegraphics[width=\linewidth]{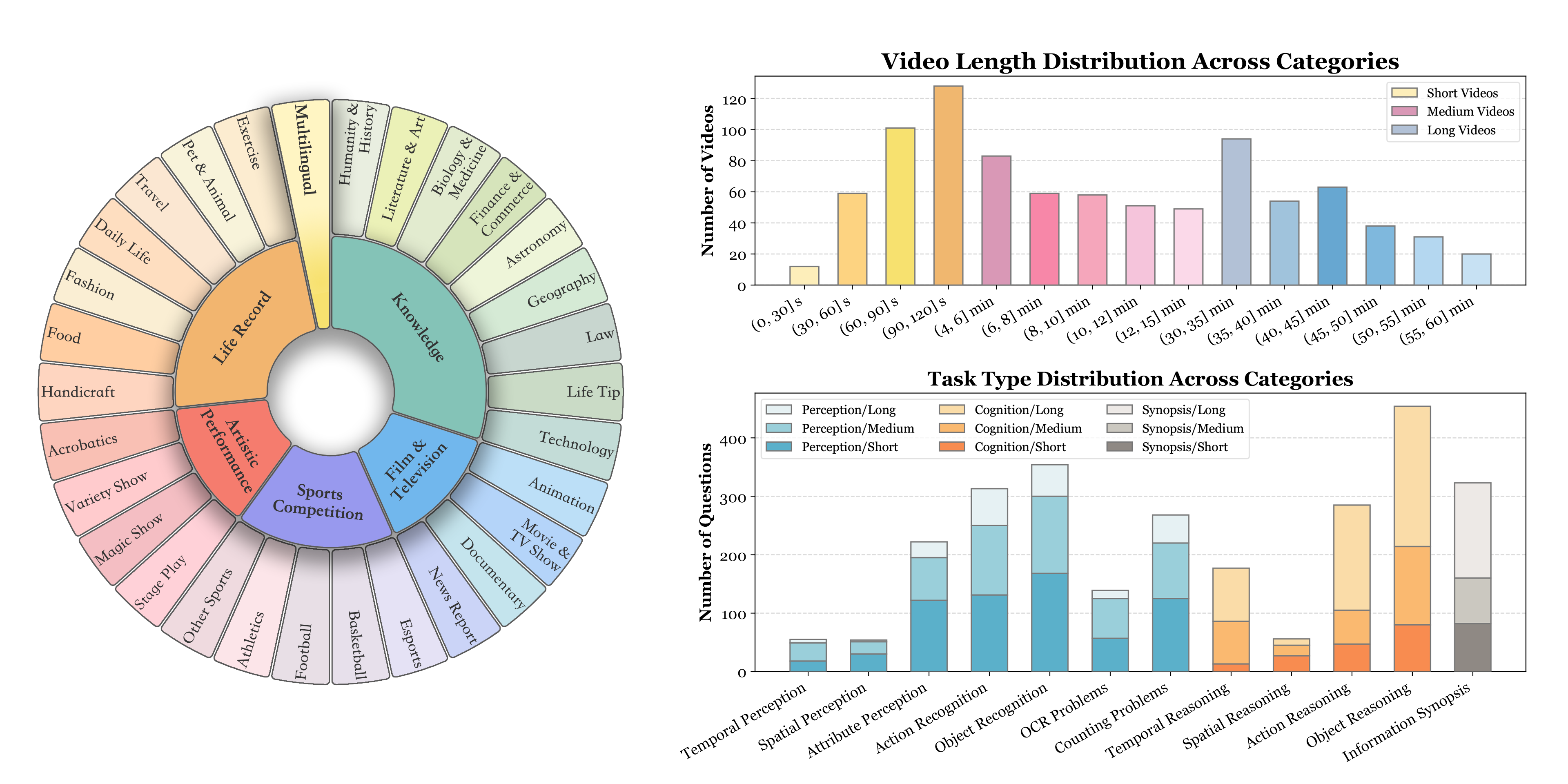}
    \caption{\textbf{Statistics analysis of Video-MME.} \textit{(left) }Video categories. Our benchmark covers 6 key domains and 30 sub-class video types. \textit{(right)} Video duration and question type distributions. Video-MME has a full spectrum of video length and covers different core abilities of MLLMs, enabling a comprehensive evaluation of temporal understanding.}
    \label{fig:domain_category}
\end{figure*}

\begin{figure}[t]
	\centering
        \begin{minipage}[h]{0.9\linewidth}
	\centering
        \scalebox{0.9}{
        \begin{tabular}{lcccc}
         \toprule 
         Dataset    & Question & Options & Answer & Subtitles \\
         \midrule
         Short & 11.5 & 17.2 & 4.0 & 198.6 \\ 
         Medium & 12.2 & 20.6 & 5.0 & 1425.6 \\
         Long & 14.5 & 31.0 & 7.5 & 6515.6\\ 
         \midrule 
         All & 12.7 & 22.9 & 5.5 & 3086.5 \\ 
         \bottomrule
        \end{tabular}}
        
        \captionof{table}{\textbf{Average word counts of different textual fields in Video-MME}. The word counts for questions, options, and answers are consistent across video lengths, indicating a standardized QA format and an increasing trend of subtitles information volume.}
        \label{tab:text_statist-cs}
	\end{minipage}%
    
	\begin{minipage}[h]{0.9\linewidth}
        \centering
            \scalebox{0.9}{\begin{tabular}{lrrr}
            \toprule
             Video     &  Avg. V.L.  & Med. C.L. & Avg. C.L. \\
             \midrule
             EgoSchema &  180 &  $\sim$ 100 &  - \\ 
            \midrule
            Short   & 82.5  &  26.0 & 28.8 \\
            Medium    &  562.7 & 164.7 & 160.0 \\
            Long   & 2385.5 & 890.7 &967.7 \\
            \bottomrule
            \end{tabular}
        }
            \captionof{table}{\textbf{Analysis of certificate length in seconds}. \textbf{Avg. V.L.}: average video length, \textbf{Med. C.L.}: median certificate length,
            \textbf{Avg. C.L.}: average certificate length.}
            \label{tab:certificate_length}
	\end{minipage}%
        \vspace{-10pt}
	
\end{figure}

\subsection{Dataset Construction}
\label{subsec:data_construction}
The construction of the Video-MME dataset involves three main steps: video collection, question-answer annotation, and quality review. The details are as follows.

\paragraph{Video Collection.}
To ensure comprehensive coverage across diverse video types, we first establish a domain hierarchy for sourcing raw videos from YouTube.
We identify 6 key domains: Knowledge, Film \& Television, Sports Competition, Life Record, and Multilingual, informed by popular trends on YouTube.
Each domain is further subdivided into specific tags, such as football and basketball under Sports Competition, yielding a total of 30 finely-grained video categories. 
The complete domain-tag hierarchy and its distribution are illustrated on the left side of Figure~\ref{fig:domain_category}.
For each category, we collect videos with varying duration lengths, including short (\textless~2 minutes), medium (4-15 minutes), and long videos (30-60 minutes). 
Besides, we also obtain corresponding meta-information such as subtitles (if provided) and audios for in-depth investigation. 
The resulting dataset consists of 900 videos, including 744 videos with subtitles and all 900 with audios, representing a range of domains with a relatively balanced distribution of video durations, as depicted on the right side of Figure~\ref{fig:domain_category}.

\paragraph{Question-Answer Annotation.}
Following the collection of raw video data, we annotate the videos with high-quality question-answer (QA) pairs to rigorously assess the proficiency of MLLMs in video content interpretation.
We employ a multiple-choice QA format to facilitate a straightforward and flexible assessment and recruit annotators with strong English proficiency and extensive research experience in vision-language learning.
Specifically, each annotator is tasked with thoroughly viewing the entire video, then iteratively creating three relevant questions, each accompanied by four candidate options, contributing to a total of 2,700 QA pairs.
As shown in the bottom right corner of Figure~\ref{fig:domain_category}, these questions encompass 12 distinct task types, including perception, reasoning, and information synthesis.
Each QA pair is designed to be closely tied to the video content, preventing MLLMs from answering accurately without directly referencing the video.

\paragraph{Quality Review.}
To guarantee the quality of our dataset, we conduct a rigorous manual review process. 
First, each QA pair undergoes examination by a different annotator to verify that (i) the language expression is clear and precise; (ii) the question is answerable with logically consistent candidate options, and the designated correct answer is appropriate.
Furthermore, to ensure that the questions are sufficiently challenging enough and require video content as a critical clue to answer~\citep{zhang2024mathverse}, we provide the text-only questions to Gemini 1.5 Pro and filter out QA pairs that can be answered solely based on the textual questions. 
For instance, a question like “What is the biggest achievement of Argentina’s number 10 in 2022?” which can be inferred as the World Cup victory, would be excluded in this phase.
Questions that do not meet this criterion are returned to the original annotator for revision. 
Statistical analysis shows that Gemini 1.5 Pro achieves less than 15\% accuracy in the text-only setup, underscoring the robustness of the video content-based requirement.
Through our dataset construction process, we strive to deliver a high-quality, diverse, and well-balanced dataset that will serve as a valuable resource for advancing research in multimodal understanding.

\subsection{Dataset Statistics}
\label{subsec:data_statistics}
In this section, we present the detailed statistics of our dataset to provide a more comprehensive understanding, including the meta information, QA pairs, certificate lengths, qualitative analysis, and comparison to previous works.

\paragraph{Video \& Meta Information.}

Our dataset comprises a total of 900 videos, 744 subtitles, and 900 audio files. 
Most videos are accompanied by both subtitles and audios, providing valuable resources for investigating the impact of external information on video understanding performance.
The upper right part of Figure~\ref{fig:domain_category} illustrates the duration distribution of the collected videos. Specifically, within the short video category, longer videos occupy a larger portion. 
For medium-length videos, the duration distribution is more uniform, while long videos exhibit a long-tailed distribution, with fewer samples as the duration increases.
The bottom right part of Figure~\ref{fig:domain_category} shows the distribution of task types. 
Shorter videos predominantly involve perception-related tasks such as action and object recognition, while longer videos mainly feature tasks related to reasoning.
Overall, this analysis highlights that Video-MME covers a wide range of video durations and task types, enabling a comprehensive evaluation of temporal understanding.

\paragraph{QA Pairs.}
We present a detailed analysis of language diversity in the questions and answers within our dataset. Table~\ref{tab:text_statist-cs} lists the average word counts of the textual fields including questions, options and answers in our dataset.
Notably, the word counts display notable consistency across different video lengths, suggesting a standardized format in the QA pairs.
In contrast, subtitle word counts increase markedly with video length—e.g., short videos average 198.6 words, while long videos reach up to 6.5K words.
This trend indicates longer videos contain more information, as evidenced by the increased volume of subtitles. 
The analysis reveals that our questions are diverse, and the answers are well-balanced. 
In addition, the distribution of the options (A/B/C/D) follows a near-uniform distribution (25.1/27.2/25.3/22.4\%), ensuring an unbiased evaluation.

\paragraph{Certificate Length Analysis.}
Inspired by the recent work EgoSchema~\citep{mangalam2024egoschema}, we adopt the \emph{certificate length} to analyze the temporal difficulty of the QA pairs.
The certificate of a given video QA pair is defined as the minimum set of sub-clips of the video that are both necessary and sufficient to convince a human verifier that the marked annotation is correct.
The certificate length is calculated as the sum of the temporal lengths of the sub-clips identified.
We randomly sample 3 videos from each class and calculate the certificate length distribution. 
As shown in Table~\ref{tab:certificate_length}, our dataset yields a median certificate length of 26s, 164.7s, and 890.7s for short, medium and long videos, respectively. Compared with the certificate length of EgoSchema, our 
medium and long video subset requires much longer video content digestion to answer the question.
To the best of our knowledge, this analysis makes our Video-MME the most challenging Video QA dataset to date.

\begin{table*}
\centering
\begin{adjustbox}{max width=0.9\linewidth}
\begin{tabular}{l!{\vrule width \lightrulewidth}c!{\vrule width \lightrulewidth}cccccccc} 
\toprule
\multicolumn{1}{c}{\multirow{2}{*}{\textbf{Models}}}              & \multirow{2}{*}{\begin{tabular}[c]{@{}c@{}}\textbf{LLM}\\\textbf{Params}~~\end{tabular}} & \multicolumn{2}{c}{\textbf{Short (\%)}} & \multicolumn{2}{c}{\textbf{Medium (\%)}} & \multicolumn{2}{c}{\textbf{Long (\%)}} & \multicolumn{2}{c}{\textbf{Overall (\%)}}  \\ 

\cmidrule(lr){3-4}\cmidrule(lr){5-6}\cmidrule(lr){7-8}\cmidrule(lr){9-10}
                                              &                                                                                     & w/o subs      & w/ subs                 & w/o subs      & w/ subs                  & w/o subs      & w/ subs                & w/o subs      & w/ subs                    \\ 
\midrule

\multicolumn{10}{c}{\textit{Open \& Closed-source Image MLLMs}}                                                                                                                                                                                                                                                          \\ 
\midrule
Qwen-VL-Chat~\citep{Qwen-VL}                                                      & 7B                                                                                       & 46.9          & 47.3                    & 38.7          & 40.4                     & 37.8          & 37.9                   & 41.1          & 41.9                       \\
Qwen-VL-Max~\citep{Qwen-VL}                                                       & -                                                                                        & 55.8          & 57.6                    & \textbf{49.2}          & 48.9                     & \textbf{48.9} & \textbf{47.0}          & \textbf{51.3} & 51.2                       \\
InternVL-Chat-V1.5~\citep{chen2024far}                                            & 20B                                                                                      & \textbf{60.2} & \textbf{61.7}           & 46.4 & \textbf{49.1}            & 45.6          & 46.6                   & 50.7          & \textbf{52.4}              \\ 
\midrule
\multicolumn{10}{c}{\textit{Open-source Video MLLMs}}                                                                                                                                                                                                                                                                                                                    \\ 
\midrule
Video-LLaVA~\citep{lin2023video}                                                  & 7B                                                                                       & 45.3          & 46.1                    & 38.0          & 40.7                     & 36.2          & 38.1                   & 39.9          & 41.6                       \\
ST-LLM~\citep{liu2023one}                                                         & 7B                                                                                       & 45.7          & 48.4                    & 36.8          & 41.4                     & 31.3          & 36.9                   & 37.9          & 42.3                       \\
ShareGPT4Video~\citep{chen2024sharegpt4video}                                                         & 8B                                                                                       & 48.3 & 	53.6 	& 36.3 	& 39.3  & 35.0  &	37.9 & 	39.9 &	43.6                    \\
VideoChat2-Mistral~\citep{li2023mvbench}                                          & 7B                                                                                       & 48.3          & 52.8                    & 37.0          & 39.4                     & 33.2          & 39.2                   & 39.5          & 43.8                       \\
Chat-UniVi-V1.5~\citep{jin2023chatunivi}                                          & 7B                                                                                       & 45.7          & 51.2                    & 40.3          & 44.6                     & 35.8          & 41.8                   & 40.6          & 45.9                       \\
VITA-1.5~\citep{fu2025vita}                                          & 7B                                                                                       & 67.0          & 69.9                    & 54.2          & 55.7                     & 47.1          & 50.4                   & 56.1          & 58.7                       \\
VITA-1.0~\citep{fu2024vita}                                          & 8$\times$7B                                                                                       & 65.9          & \textbf{70.4}                    & 52.9          & 56.2                     & 48.6          & 50.9                   & 55.8          & 59.2                       \\
LLaVA-NeXT-Video~\citep{zhang2024llavanextvideo}                                  & 34B                                                                                      & 61.7          & 65.1                    & 50.1          & 52.2                     & 44.3          & 47.2                   & 52.0          & 54.9                       \\
VILA-1.5~\citep{lin2023vila}                                &  34B & \textbf{68.1} & 68.9           & \textbf{58.1} & \textbf{57.4}            & \textbf{50.8} & \textbf{52.0}          & \textbf{59.0} & \textbf{59.4}              \\ 
\midrule
\multicolumn{10}{c}{\textit{Closed-source MLLMs}}                                                                                                                                                                                                                                                                                                                        \\ 
\midrule
GPT-4V~\citep{openai2023gpt4v} & -                                                                                        & 70.5          & 73.2                    & 55.8          & 59.7                     & 53.5          & 56.9                   & 59.9          & 63.3                       \\
GPT-4o~\citep{openai2024gpt4o}                                                    & -                                                                                        & 80.0          & 82.8                    & 70.3          & 76.6                     & 65.3          & 72.1                   & 71.9          & 77.2                       \\
Gemini 1.5 Flash~\citep{Reid2024Gemini1U}                              &  - & 78.8 	& 79.8 	& 68.8 	& 74.7 &	61.1 &	68.8 &	70.3 	& 75.0                             \\
Gemini 1.5 Pro~\citep{Reid2024Gemini1U}                                           & -                                                                                        & \textbf{81.7} & \textbf{84.5}           & \textbf{74.3} & \textbf{81.0}            & \textbf{67.4} & \textbf{77.4}          & \textbf{75.0} & \textbf{81.3}              \\
\bottomrule

\end{tabular}
\end{adjustbox}
\caption{\textbf{Performance of MLLMs on Video-MME.} The results are evaluated on short, medium, and long durations, under the settings of ``without subtitles'' and ``with subtitles''. In the “with subtitles” setting, subtitles are selected to correspond to the sampled video frames.}
\label{tab:open-source-eval-results-wrt-duration}
\vspace{-1em}
\end{table*}

\paragraph{Qualitative Analysis.}
\label{para:qualitative-analysis}
Building on our previous analysis, we have established that our proposed benchmark, Video-MME, is both diverse and challenging, making it an exemplary testbed for MLLMs. Figure~\ref{fig:video-mme-case} showcases specific cases from our Video-MME dataset to illustrate this.

In the first example, the model must integrate information from various sources: visual data from video frames (e.g., ``Day 1 is May 31, 2021'') and auditory/subtitle content (e.g., referring to ``Yosemite National Park''). Moreover, the model is required to perform simple arithmetic operations to determine the exact departure date. This multi-modal and multi-step reasoning highlights the complexity and high quality of our dataset.
The second example involves a question placed towards the end of a video, with the provided answer options dispersed across different segments of the video. This necessitates a comprehensive understanding of the entire content, which can be as long as 30 minutes. 
These highlighted cases underscore that our Video-MME dataset is meticulously designed to pose significant challenges, thereby effectively evaluating the compositional video understanding capabilities of MLLMs.


\section{Experiments}
In this section, we evaluate a variety of MLLMs on our Video-MME benchmark.
We begin by outlining the evaluation settings, followed by the quantitative results for both open-source and closed-source models
Finally, we present case studies to provide an intuitive understanding, and investigate the effect of the modality and duration.

\begin{figure}
    \centering
    \includegraphics[width=0.9\linewidth]{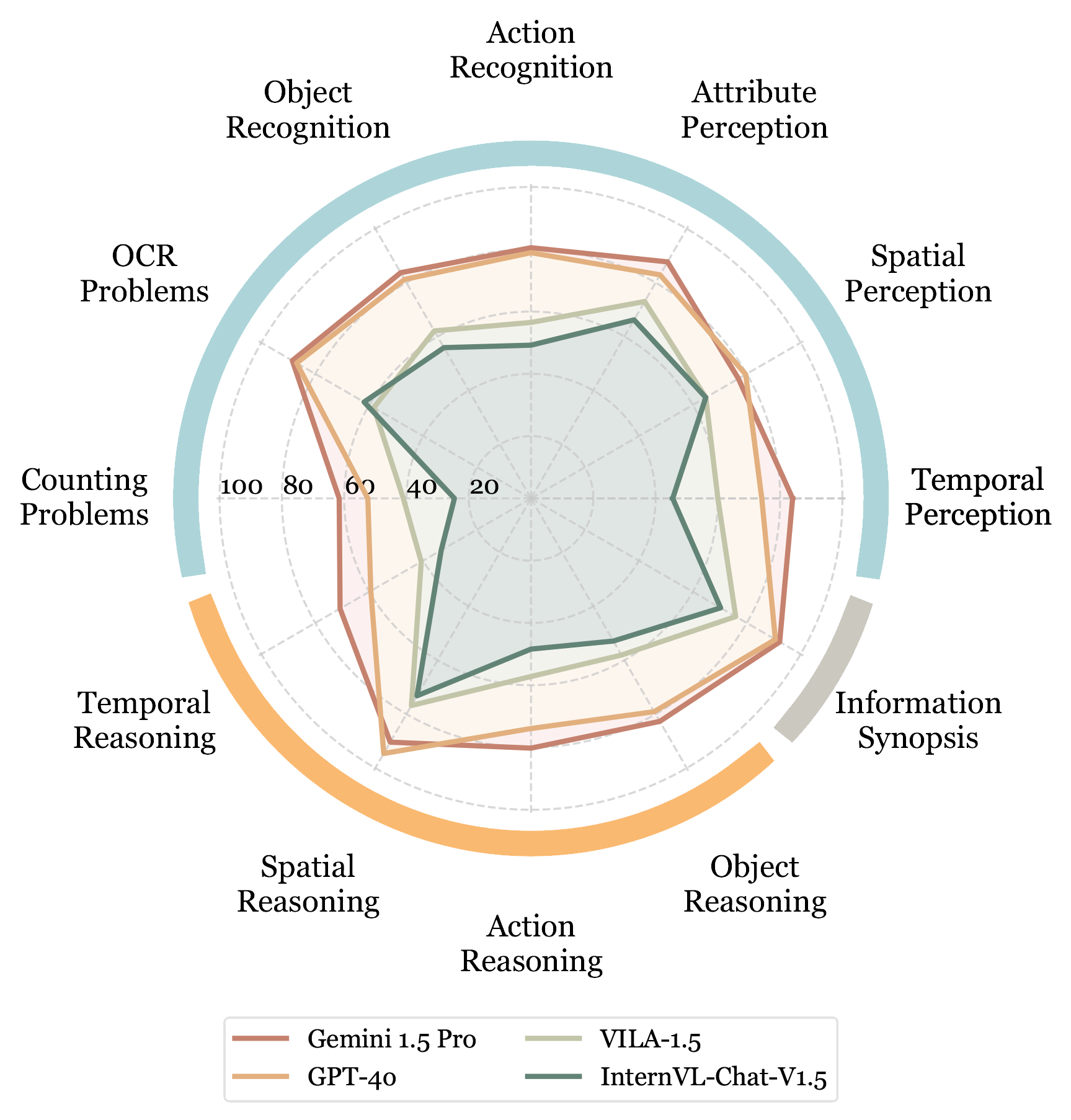}
    \caption{\textbf{Comparison between four selected MLLMs on Video-MME across different tasks.} The \textcolor{my_light_blue}{light blue} in the outer circle stands for perception tasks and the \textcolor{my_orange}{orange} represents reasoning tasks. As can be observed, counting problems are a joint bottleneck of current
multimodal models in video understanding.}
    \label{fig:gemini-eval-wrt-question-type}
    \vspace{-1em}
\end{figure}

\begin{table*}
\centering
\begin{adjustbox}{max width=0.85\linewidth}
\begin{tabular}{crlllllll} 
\toprule
\multirow{3}{*}{\textbf{Subset}} & \multicolumn{1}{c}{\multirow{3}{*}{\textbf{Modality}}} & \multicolumn{7}{c}{\textbf{Category }}                                                                                                                                                                                                                                                                                                                                                \\ 
\cmidrule{3-9}
                                 & \multicolumn{1}{c}{}                               & \multicolumn{1}{c}{\multirow{2}{*}{\textit{Knowledge~}}} & \multicolumn{1}{c}{\multirow{2}{*}{\textit{Film \& Television}}} & \multicolumn{1}{c}{\textit{Sports}}      & \multicolumn{1}{c}{\textit{Artistic~}}   & \multicolumn{1}{c}{\textit{Life}}   & \multicolumn{1}{c}{\multirow{2}{*}{\textit{Multilingual }}} & \multicolumn{1}{c}{\multirow{2}{*}{Overall}}  \\
                                 & \multicolumn{1}{c}{}                               & \multicolumn{1}{c}{}                                     & \multicolumn{1}{c}{}                                                             & \multicolumn{1}{c}{\textit{Competition}} & \multicolumn{1}{c}{\textit{Performance}} & \multicolumn{1}{c}{\textit{Record}} & \multicolumn{1}{c}{}                                        & \multicolumn{1}{c}{}                          \\ 
\midrule
\multirow{3}{*}{Short} & Frames & 78.3 & 80.8 & 76.7 & 86.7 & 88.1 & 76.7 & \textbf{81.7} \\
 & + Subs & 83.3 \textit{(+4.9)} & 86.7 \textit{(+5.8)} & 79.3 \textit{(+2.7)} & 87.6 \textit{(+1.0)} & 86.6 \textit{(-1.5)} & 86.7 \textit{(+10.0)} & \textbf{84.5} \textit{(+2.8)} \\
 & + Audio & 81.4 \textit{(+3.1)} & 87.5 \textit{(+6.7)} & 78.7 \textit{(+2.0)} & 86.7 \textit{(-)} & 85.6 \textit{(-2.4)} & 86.7 \textit{(+10.0)} & \textbf{83.6} \textit{(+1.9)} \\
\midrule
\multirow{3}{*}{Medium} & Frames & 70.2 & 81.2 & 68.7 & 84.3 & 73.4 & 87.5 & \textbf{74.3} \\
 & + Subs & 83.3 \textit{(+13.1)} & 84.9 \textit{(+3.7)} & 76.2 \textit{(+7.5)} & 85.3 \textit{(+1.0)} & 76.8 \textit{(+3.4)} & 83.3 \textit{(-4.2)} & \textbf{81.0} \textit{(+6.7)} \\
 & + Audio & 80.2 \textit{(+9.9)} & 83.9 \textit{(+2.6)} & 72.1 \textit{(+3.4)} & 84.3 \textit{(-)} & 76.8 \textit{(+3.4)} & 100.0 \textit{(+12.5)} & \textbf{79.5} \textit{(+5.2)} \\
\midrule
\multirow{3}{*}{Long} & Frames & 73.4 & 70.1 & 58.3 & 63.3 & 65.1 & 70.8 & \textbf{67.4} \\
 & + Subs & 83.0 \textit{(+9.6)} & 71.4 \textit{(+1.3)} & 77.5 \textit{(+19.2)} & 70.0 \textit{(+6.7)} & 74.6 \textit{(+9.5)} & 87.5 \textit{(+16.7)} & \textbf{77.4} \textit{(+10.1)} \\
 & + Audio & 81.1 \textit{(+7.7)} & 73.2 \textit{(+3.1)} & 72.6 \textit{(+14.3)} & 63.3 \textit{(-)} & 66.7 \textit{(+1.6)} & 83.3 \textit{(+12.5)} & \textbf{73.6} \textit{(+6.2)} \\
\midrule
\multirow{3}{*}{Overall} & Frames & 74.1 & 77.9 & 68.6 & 78.8 & 77.4 & 78.2 & \textbf{75.0} \\
 & + Subs & 83.2 \textit{(+9.2)} & 81.8 \textit{(+3.9)} & 77.7 \textit{(+9.1)} & 81.5 \textit{(+2.7)} & 80.3 \textit{(+2.9)} & 85.9 \textit{(+7.7)} & \textbf{81.3} \textit{(+6.2)} \\
 & + Audio & 80.9 \textit{(+6.8)} & 82.4 \textit{(+4.5)} & 74.6 \textit{(+6.1)} & 78.8 \textit{(-)} & 78.0 \textit{(+0.6)} & 89.7 \textit{(+11.5)} & \textbf{79.4} \textit{(+4.3)} \\
\bottomrule
\end{tabular}
\end{adjustbox}
\caption{\textbf{Performance of Gemini 1.5 Pro across six major categories in the Video-MME benchmark.} Evaluation includes three input modality settings: frames only, frames with subtitles, and frames with audio, highlighting the impact of multimodal inputs on model performance within each category. As one can observe, subtitles and audios contribute essential information for understanding the video.}

\label{gemini-eval-results-wrt-duration-topic}
\vspace{-1em}
\end{table*}

\subsection{Settings}
\label{subsec:setting}

We conduct the evaluation on 4 commercial models, i.e., GPT-4V~\citep{openai2023gpt4v}, GPT-4o~\citep{openai2024gpt4o}, Gemini 1.5 Flash~\citep{Reid2024Gemini1U}, and Gemini 1.5 Pro~\citep{Reid2024Gemini1U}. Representative open-source video MLLMs including Video-LLaVA~\citep{lin2023video}, VideoChat2-Mistral\citep{li2023mvbench}, ST-LLM~\citep{liu2023one}, ShareGPT4Video~\citep{chen2024sharegpt4video},Chat-UniVi-V1.5~\citep{jin2023chatunivi}, LLaVA-NeXT-Video~\citep{zhang2024llavanextvideo}, VITA-1.0~\citep{fu2024vita}, VITA-1.5~\citep{fu2025vita}, and VILA-1.5~\citep{lin2023vila} are evaluated as well. 
In addition, we also include advanced image MLLMs, i.e., Qwen-VL-Chat/Max~\citep{Qwen-VL} and InternVL-Chat-V1.5~\citep{chen2024far}, which usually can generalize to multi-image scenarios.
We follow their official configurations and try to use more frames for evaluation. 
The accuracy is computed by directly comparing the model’s output with the ground-truth answer, without utilizing any external models, such as ChatGPT.

\subsection{Quantitative Results}
\label{subsec:metrics}

\paragraph{Performance of Commercial Models.} 
As one of the pioneering commercial large models integrated with video comprehension capabilities, Gemini 1.5 Pro has achieved the best performance among its peers on \datasetname. 
As depicted in Table~\ref{tab:open-source-eval-results-wrt-duration}, with video frames as input alone, Gemini 1.5 Pro attains an accuracy of 75\%, surpassing GPT-4V and GPT-4o by 15.1\% and 3.1\%, respectively. 
Table~\ref{gemini-eval-results-wrt-duration-topic} shows the fine-grained performance of Gemini 1.5 Pro. 
Among the 6 major video categories, Gemini 1.5 Pro performs the best in Artistic Performance while performing the lowest in the Sports Competition category.
As video duration increases, Gemini 1.5 Pro's performance declines (e.g., $-$14.3\% from short to long videos), highlighting the model's weakness in capturing long-range temporal relationships. 
Nevertheless, Gemini 1.5 Pro's performance on long videos still surpasses almost all open-source models, except for VILA-1.5, on short videos, demonstrating its superior capabilities. 
In addition to visual frame input, Gemini 1.5 Pro's support for additional modalities, including subtitles and audios, provides opportunities for further performance improvement. 
For example, Table~\ref{gemini-eval-results-wrt-duration-topic} displays that using audios can increase accuracy by 6.2\% for long videos, and the improvement in the multilingual category even reaches 16.7\%. 
We can also see that the effect of subtitles and audios is different in these six categories.
These motivate future research to develop versatile models that can support a wider range of modality inputs.

\paragraph{Performance of Open-sourced Models.}

As shown in Table~\ref{tab:open-source-eval-results-wrt-duration}, among the 7B models, VITA-1.5 achieves the best performance with 56.1\%.  
VILA-1.5 with 34B LLM achieves an accuracy of 59\%, demonstrating its stronger capabilities, especially in the tasks of spatial reasoning, attribute perception, and information synopsis (Figure~\ref{fig:gemini-eval-wrt-question-type}).
Nevertheless, there remains a significant gap between VILA-1.5 and Gemini Pro 1.5, particularly in counting problems, action recognition, and temporal perception, indicating substantial room for improvement. 
Regarding modality, adding subtitles consistently improves the performance of open-source models. Unfortunately, most of the open-source models do not support audio input, so audio evaluation is omitted.

Apart from video MLLMs, we also evaluate the performance of image MLLMs on Video-MME.
Table~\ref{tab:open-source-eval-results-wrt-duration} reveals that image-based Qwen-VL-Max and InterVL-Chat-V1.5 attain comparable performance to LLaVA-NeXT-Video, demonstrating their superior generalization capacity on sequential data, and the universality of Video-MME in both image and video MLLMs.
It also indicates that image understanding is the foundation of video understanding.

\subsection{Analysis}
\label{subsec:exp_analysis}
We conduct further analysis to explore the factors influencing the video understanding performance, e.g., additional modality information and video duration.

\paragraph{\textit{Could additional modalities benefit the performance?}}
Most evaluations use only video frames as input, requiring models to answer questions based solely on visual context.
However, many videos inherently include additional information from other modalities, such as subtitles and audios.
To understand their impact, we vary the combinations of input modalities in the evaluation.
We can draw the following observations from the results in Table~\ref{gemini-eval-results-wrt-duration-topic}. 
\textbf{(1) Introducing subtitles and audios can improve the results.} 
For instance, in the multilingual task presented in Table~\ref{gemini-eval-results-wrt-duration-topic}, Gemini 1.5 Pro achieves a +16.7\% and +12.5\% accuracy improvement on long videos with the addition of subtitles and audios, respectively, compared to the frame-only setting. This suggests that subtitles and audios contribute essential information for answering the questions.
\textbf{(2) Subtitles and audios provide greater assistance in understanding long videos compared to short videos.} 
For example, in Table~\ref{gemini-eval-results-wrt-duration-topic}, compared to only using frames, the addition of subtitles improves the model's performance by 2.8\% on short videos and by 10.1\% on long videos. 
This is because test samples of long videos include more challenging reasoning questions, which require the model to utilize subtitles and audio information for accurate responses. 
\textbf{(3) For MLLMs, using subtitles is more effective than audios.} 
Subtitles are typically transcriptions of speech, focusing on verbal content, while audio includes additional ambient sounds.
As shown in Table~\ref{gemini-eval-results-wrt-duration-topic}, subtitles consistently provide greater improvement than audio across different durations of videos.
Multilingual can be regarded as an exception, possibly due to the quality of the subtitles.

\paragraph{\textit{How MLLMs are robust to varied video duration?}}
In Table~\ref{tab:open-source-eval-results-wrt-duration}, we respectively compare the performance of different models on short, medium, and long videos. As video duration increases, both open-sourced and commercial models exhibit a significant decline in performance. 
There are three main reasons for the performance decline. 
\textbf{(1) Increased proportion of difficult tasks.} 
As shown in the bottom right corner of Figure~\ref{fig:domain_category}, test samples for long videos contain a higher proportion of reasoning questions, which poses a greater challenge to the model's capabilities. 
\textbf{(2) Increased sparsity in frame sampling, leading to a reduction in effective input information.} 
Ideally, for videos of varying lengths, models should sample video frames at a fixed fps to ensure consistent information density in frame sequences~\citep{timechat, testa}. 
However, existing open-sourced models fix the number of input frames, e.g., 8 frames, resulting in excessively sparse information density as the video length increases.
This sparsity prevents the model from retaining all useful visual semantics, hindering accurate predictions. 
Introducing additional modalities, e.g., subtitles, can effectively supplement the missing information~\citep{Chen2024TowardsMV}.
\textbf{(3) Increased difficulty in long context understanding.}
Although Gemini 1.5 Pro correspondingly increases the number of sampled frames in the long video, there is still a significant performance degradation.
Understanding the long context of either single-modality (LLM) or multi-modality (MLLM) is always a great challenge.

\section{Discussions}
Our evaluation using Video-MME has revealed several critical insights into the current MLLMs and highlighted areas for future improvement. 
In this section, we provide more discussions on potential future directions.

\paragraph{\textit{Improving Long Context Modeling Capabilities of MLLMs.}}
One of the significant challenges identified in our evaluation is the decline in performance as video duration increases.
For open-source models, the restricted input frames can become an information bottleneck for understanding the full content of long videos, which advocates for innovative approaches of both architectural and infrastructural context extension.
For instance, exploring techniques like ring attention~\citep{liu2023ring}, as investigated by large world models~\citep{liu2023world}, and training-free context extension methods could be beneficial~\citep{an2024chunkllama}.
Additionally, developing architectures such as a temporal Q-Former to adaptively identify key frames in the video or compress video tokens to reduce computational overhead based on the questions posed is also worth exploring~\citep{timechat,dai2023instructblip}. 
In essence, improving long context modeling ability is crucial for the next generation of MLLMs to understand long sequential world dynamics.

\paragraph{\textit{Building Datasets with Complex Temporal Understanding.}}
Our evaluation emphasizes the need for instruction-tuning datasets focused on temporal reasoning, particularly given the prevalence of traditional video datasets with short inputs, such as MSRVTT-QA and ActivityNet-QA.
Although there have been efforts to construct high-quality datasets involving complex temporal reasoning in long videos~\citep{mangalam2024egoschema,li2023videochat}, the availability of such datasets remains limited compared to those for text~\citep{longpre2023flan,mishra2022naturalinstruction} and image~\citep{li2023m3it,visionFlan2023}. 
The long-tailed distribution of this data poses challenges for acquisition.
Advancements in annotation methods, such as human-in-the-loop frameworks~\citep{li2023vitatecs} and explorations into automatic data synthesis are crucial~\citep{liu2024best}. 
Developing these datasets will enable better use of architectural innovations, providing MLLMs with sufficient training supervision for robust temporal understanding in videos.

\section{Conclusion}
In this paper, we have introduced Video-MME, which is designed to evaluate MLLMs on video understanding tasks. 
Our benchmark incorporates a diverse range of video types, temporal durations, and data modalities with high-quality, expert-annotated QA pairs.
Our extensive evaluation underscores the need for further advancements in handling longer multimodal data and we hope Video-MME will inspire future research and development in improving the capabilities of MLLMs.

\section*{Acknowledgments}
This work was funded by National Natural Science Foundation of China under Grant 62441234. 
We appreciate the efforts made by Yu Bai, Fangyuan Liu, Yigeng Jiang, and Zezhong Wu in the construction of the benchmark.

{
    \small
    \bibliographystyle{ieeenat_fullname}
    \bibliography{main}
}

\clearpage
\setcounter{page}{1}
\maketitlesupplementary

\section{Detailed Experimental Settings}

\paragraph{Models.}
We conduct a comprehensive evaluation on four commercial models and nine representative open-source video-based multimodal large language models. To further demonstrate the adaptability of our benchmark to multi-image scenarios, we also include three widely utilized image-based MLLMs as part of the evaluation. The complete list of models evaluated is provided below.
\begin{itemize}
    \item \textbf{Commercial MLLMs}: GPT-4V, GPT-4o, Gemini 1.5 Flash, and Gemini 1.5 Pro.
    \item \textbf{Open-source Video MLLMs}: Video-LLaVA, ST-LLM, ShareGPT4Video, VideoChat2-Mistral, VILA-1.5, Chat-UniVi-V1.5, VITA-1.0, VITA-1.5, LLaVA-NeXT-Video.
    \item \textbf{Advanced Image MLLMs}: Qwen-VL-Chat/Max and InternVL-Chat-V1.5.
\end{itemize}

\paragraph{Frame Extraction.}
A standard approach involves extracting a sequence of frames from the video and interpreting the resulting multi-image inputs. 
For Gemini 1.5 Pro which supports extremely long multimodal contexts, we sample frames at 1 frame per second for short and medium videos, and at 1 frame every 2 seconds for long videos to ensure API stability.
For all other models, frame extraction adheres to their respective official guidelines, uniformly sampling a specified number of frames from the video.
The specific numbers of sampled frames are as follows: 10 frames for GPT-4V, 384 for GPT-4o, 8 for Video-LLaVA, 16 for VideoChat2-Mistral, 16 for ShareGPT4Video, 64 for ST-LLM, 64 for Chat-UniVi-V1.5, 32 for VITA-1.0, 16 for VITA-1.5, 32 for LLaVA-NeXT-Video, 8 for VILA-1.5, 4 for both Qwen-VL-Chat and Qwen-VL-Max,  10 for InternVL-Chat-V1.5. 
\paragraph{Subtitle Utilization.}
In the subtitle-enabled setting, all models utilize subtitles corresponding to the timestamps of the sampled video frames. 
For instance, if 10 frames are sampled from a video, the 10 subtitles that correspond to the respective timestamps of those frames are selected.
This approach ensures accurate coherent synchronization between the visual and textual multimodal input for evaluation.

\paragraph{Evaluation.}
The evaluation follows the format: “entire video frames + complete subtitles/audios (optional) + question with prompt.” Whenever possible, the model’s default prompt is utilized for multiple-choice questions. If unavailable, a standardized prompt is employed as follows:
\begin{displayquote}
\textit{This video's subtitles are listed below: [Subtitles] Select the best answer to the following multiple-choice question based on the video. Respond with only the letter (A, B, C, or D) of the correct option. [Question] The best answer is:}
\end{displayquote}
The accuracy is computed by extracting the model’s output using regular expressions and comparing it directly with the ground-truth answer, without relying on external judges like commonly-used ChatGPT.

\section{Additional Analysis}

\paragraph{\textit{How do MLLMs perform on the two highlighted cases in Figure~\ref{fig:video-mme-case}?}}
We conduct qualitative evaluation (using frames and subtitles) on the two cases in Figure~\ref{fig:video-mme-case}. 
As analyzed in Section~\ref{para:qualitative-analysis}, these two cases comprehensively examine the model's capabilities in OCR, attribute perception, object recognition, and long-range temporal reasoning, making them highly challenging. 
\textbf{For the date-related question in Case 1}, Video-LLaVA identifies the date (May 31st) from the frame at 01:10 and subtitles, but fails to perform reasoning based on context and incorrectly determines the year of the event, leading to the erroneous selection of option A. The remaining open-sourced models miscalculate the date 10 days after May 31st during the reasoning process, resulting in the incorrect choice of option C. 
\textbf{For the event-related question in Case 2}, Video-LLaVA, VideoChat2, and ST-LLM incorrectly associate the target person with nearby events, resulting in the selection of incorrect options A or C. 
In contrast, LLaVA-NeXT-Video and Gemini 1.5 Pro accurately track the events involving the target individual across the entire video, showcasing robust long-range temporal modeling capabilities. 
They correctly link the target person’s injury at 03:35 with his reappearance at 27:30, identifying the true cause of the injury (option D).
In summary, the questions in our benchmark pose significant challenges to the models, which motivates MLLMs to advance both their perception and reasoning capabilities.

\begin{figure*}
    \centering
    \includegraphics[width=0.88\linewidth]{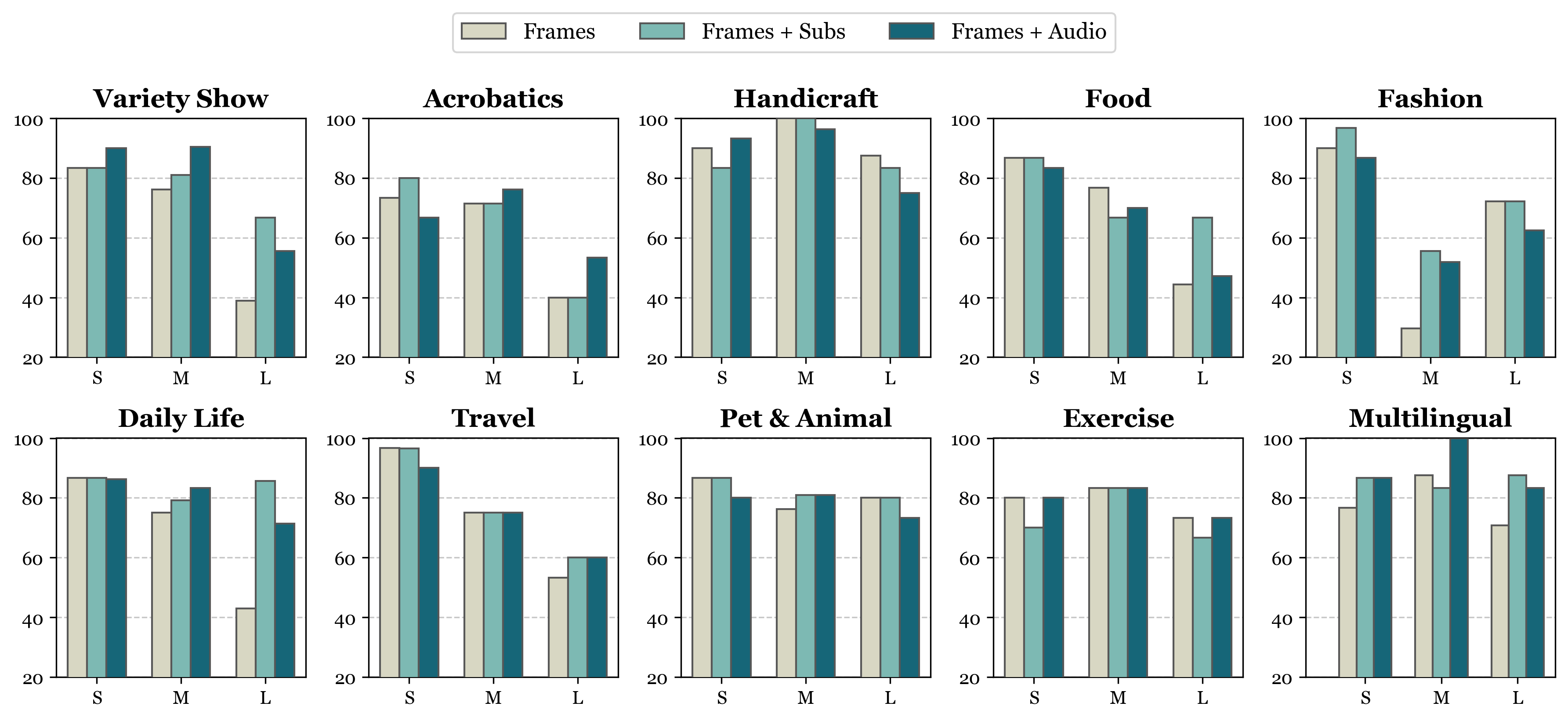}
    \caption{Evaluation results of Gemini 1.5 Pro across different video subcategories.}
    \label{fig:results_of_video_sub_types-3}
\end{figure*}

\paragraph{\textit{Could additional modalities benefit the performance across categories?}}
Figure~\ref{fig:results_of_video_sub_types-3} presents the results of Gemini 1.5 Pro across the 30 subcategories of Video-MME, under the testing modes of frames, frames + subtitles, and frames + audio. The results indicate that subtitles and audio positively contribute to video understanding in multimodal large models. However, the extent of improvement provided by these modalities varies across different domains.

\begin{figure*}
    \centering
    \includegraphics[width=0.88\linewidth]{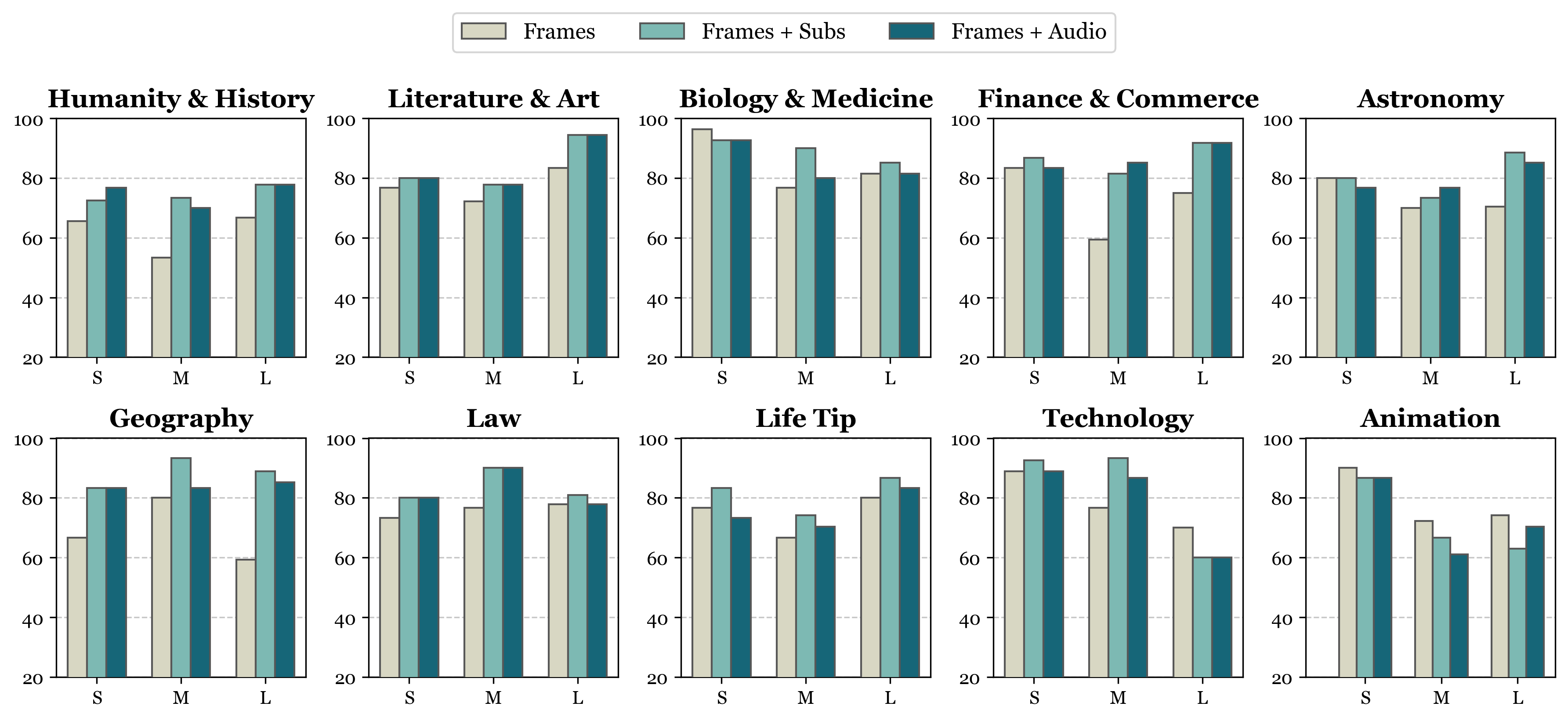}
    \label{fig:results_of_video_sub_types-1}
\end{figure*}

\begin{figure*}
    \centering
    \includegraphics[width=0.88\linewidth]{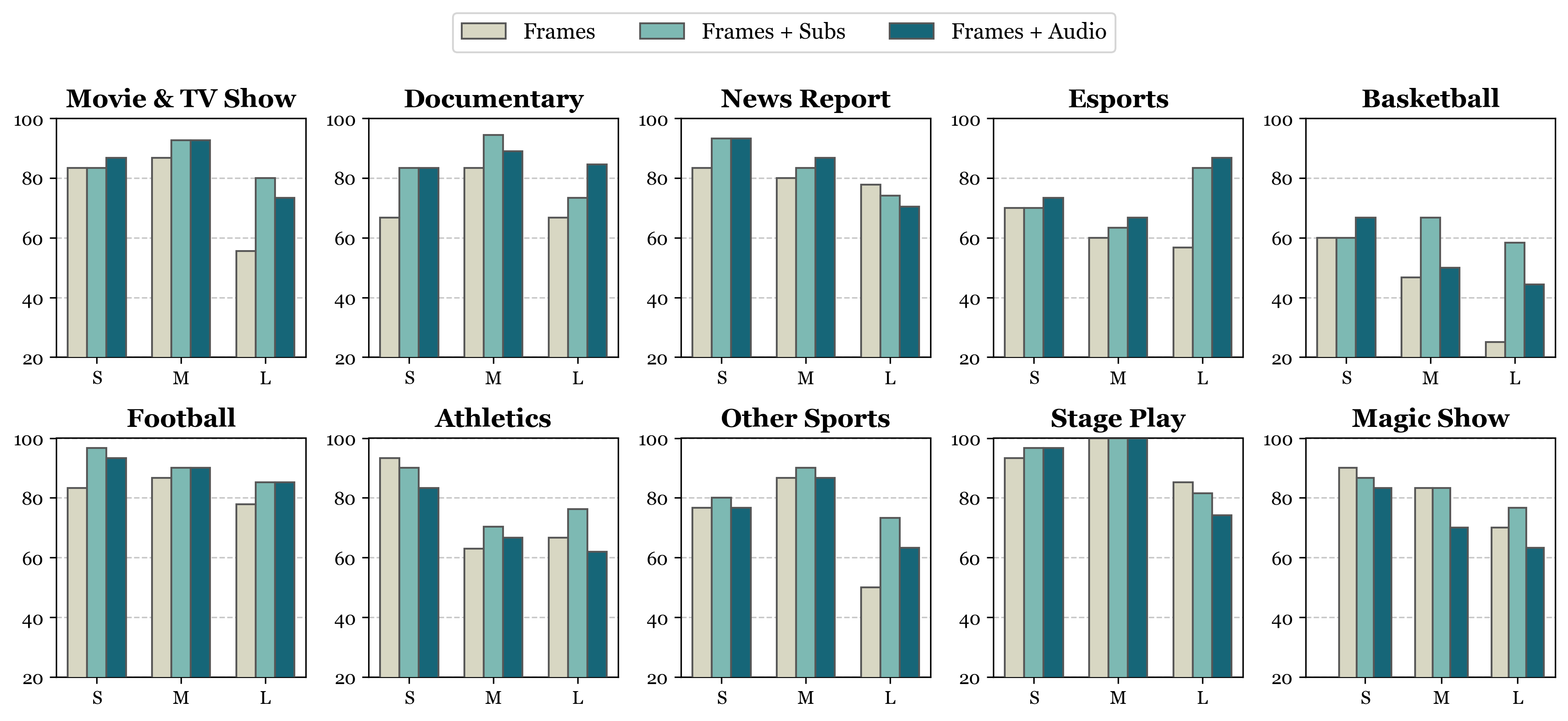}
    \label{fig:results_of_video_sub_types-2}
\end{figure*}

\end{document}